# Review of control algorithms for mobile robotics

# Revisión de algoritmos de control para robótica móvil

Andrés-David Suárez-Gómez,[1] ANDRES A. HERNANDEZ ORTEGA,[1,2]

[1] UNIVERSIDAD NACIONAL ABIERTA Y A DISTANCIA – Grupo de Investigación Byte in Design
[2] UNIVERSIDAD PEDAGOGICA Y TECNOLOGICA DE COLOMBIA.

## Abstract

This article presents a comprehensive review of control algorithms used in mobile robotics, a field in constant evolution. Mobile robotics has seen significant advances in recent years, driven by the demand for applications in various sectors, such as industrial automation, space exploration, and medical care. The review focuses on control algorithms that address specific challenges in navigation, localization, mapping, and path planning in changing and unknown environments. Classical approaches, such as PID control and methods based on classical control theory, as well as modern techniques, including deep learning and model-based planning, are discussed in detail. In addition, practical applications and remaining challenges in implementing these algorithms in real-world mobile robots are highlighted. Ultimately, this review provides a comprehensive overview of the diversity and complexity of control algorithms in mobile robotics, helping researchers and practitioners to better understand the options available to address specific problems in this exciting area of study.

## Resumen

Este artículo presenta una revisión exhaustiva de los algoritmos de control utilizados en la robótica móvil, un campo en constante evolución. La robótica móvil ha experimentado avances significativos en los últimos años, impulsados por la demanda de aplicaciones en diversos sectores, como la automatización industrial, la exploración espacial y la atención médica. La revisión se centra en algoritmos de control que abordan desafíos específicos en la navegación, localización, mapeo y planificación de trayectorias en entornos cambiantes y desconocidos. Se discuten en detalle los enfoques clásicos, como el control PID y los métodos basados en teoría de control clásica, así como las técnicas modernas, incluyendo el aprendizaje profundo y la planificación basada en modelos. Además, se destacan las aplicaciones prácticas y los desafíos pendientes en la implementación de estos algoritmos en robots móviles del mundo real. En última instancia, esta revisión proporciona una visión general integral de la diversidad y complejidad de los algoritmos de control en la robótica móvil, ayudando a los investigadores y profesionales a comprender mejor las opciones disponibles para abordar problemas específicos en esta emocionante área de estudio.

## Introducción

A menudo se investiga el diseño y la optimización de algoritmos de control para mejorar diferentes aspectos del rendimiento de un robot, como el control de seguimiento de trayectorias para moverse de un punto a otro. Hay una gran cantidad de investigaciones sobre algoritmos de control de robots y se proponen constantemente nuevos enfoques. Sin embargo, surge el problema de la dificultad para comparar los resultados de estas investigaciones publicadas y evaluar la calidad de los estudios. Además, en las publicaciones sobre robótica, los criterios de evaluación del rendimiento suelen ser pasados por alto, lo que dificulta realizar una comparación objetiva de los algoritmos. Las pruebas, ya sean en simulación o experimentales, a menudo se limitan a medir la longitud de la trayectoria recorrida o el tiempo que tarda el robot en completar una tarea. Además, hay pocos métodos estándar para evaluar las capacidades y limitaciones de estos sistemas de manera comparable. (Norton et al, 2019).

Las investigaciones en este campo se suelen llevar a cabo en entornos de laboratorio controlados para validar pruebas de concepto y establecer comparativas útiles. Sin embargo, es importante tener en cuenta que los resultados obtenidos pueden diferir en cierta medida de la operación real de un robot, ya que esta última está caracterizada por la presencia de incertidumbre (Martins et al., 2020). Además, existe una falta de consenso en cuanto a los criterios de evaluación del rendimiento, los cuales suelen variar de un estudio a otro. Esta falta de consenso dificulta la comparación de las capacidades de los algoritmos de navegación y resta rigor a la evaluación de los avances en este campo. En consecuencia, se carece de un sistema integral de evaluación (Ren et al., 2020).

## Materiales y Métodos

Normalmente el control de las entradas de un modelo monociclo trata de aplicar el controlador PID de retroalimentación tradicional y seleccionar la entrada adecuada, u = $(v\omega)^T$, dado por la ecuación:

$$U(t) = PID(e) = K_p e(t) + K_I \int_0^1 e(\tau)\, d(\tau) + K_D \frac{de\,(t)}{dt} \qquad (1)$$

En el contexto de cada tarea que se detalla a continuación, el término 'e' se refiere al error entre el valor deseado y el valor obtenido como resultado. Las constantes Kp, Kl y KD representan las ganancias proporcional, integradora y derivativa respectivamente, mientras que 't' hace referencia al tiempo. Las ganancias de control empleadas en este estudio se determinan mediante ajustes de diferentes valores con el objetivo de lograr respuestas satisfactorias. Si el vehículo se desplaza a una velocidad constante, v = v0, entonces la entrada de control solo cambiará junto con la velocidad angular, ω, de esta manera:

$$w = PID(e) \qquad (2)$$

## Resultados y Discusión

*Criterios De Espacio-Tiempo*

Los criterios de desempeño relacionados con las dimensiones espacio-tiempo son utilizados ampliamente y permiten una evaluación y comparación cuantitativa de los resultados en experimentos reales o en simulación. En el artículo de Munoz et al. (2014), se describen varios criterios típicos en navegación y evasión de obstáculos, como el éxito de la misión, la robustez en espacios estrechos, la longitud del camino, el tiempo requerido para completar la misión, los periodos de control, la distancia media al objetivo, la distancia a los obstáculos y la suavidad de la trayectoria, entre otros.

Entre estos criterios, los relacionados con las dimensiones espacio-tiempo son los más simples y comúnmente utilizados. Se considera que una trayectoria óptima desde el punto de vista de alcanzar la meta es aquella que sigue una línea recta de la mínima longitud posible y sin curvatura entre el punto de inicio (xi, yi) y el punto de llegada (xn, yn), recorrida en el menor tiempo. Este enfoque asume la linealidad y la velocidad constante del robot en su trayectoria hacia la meta (Munoz-Ceballos, N. D. et al., 2022).

El "Dynamic Window Approach" (DWA) es un conocido algoritmo de navegación para evitar colisiones. Fue propuesto inicialmente por Dieter Fox y su equipo. DWA funciona en tiempo real y reacciona a las situaciones cambiantes a medida que ocurren. En los últimos años, la función de coste de DWA ha experimentado varias extensiones y mejoras. Este enfoque determina las velocidades de translación (v) y rotación (w) seguras y óptimas directamente creando perfiles de velocidad que tienen en cuenta la dinámica del robot y las limitaciones de velocidad y aceleración. La búsqueda de velocidades adecuadas implica principalmente tres subespacios, incluyendo el espacio de posibles valores de v (Mohammadpour et al, 2021).
El espacio de posibles velocidades se divide en tres subespacios de acuerdo con las restricciones cinemáticas del robot:

Espacio de Velocidades Posibles (Vs): Este subespacio considera las limitaciones kinemáticas del robot y representa todas las velocidades que son físicamente posibles para el robot en función de sus características mecánicas.

Espacio de Velocidades Admisibles (Va): En este subespacio se contemplan las velocidades que permiten al robot detenerse sin colisionar con un obstáculo. Estas velocidades están restringidas por la capacidad del robot para detenerse de manera segura y sin colisiones.

Espacio de Velocidades Posibles con Consideración de las Limitaciones de Aceleración del Robot (Vd): Este subespacio tiene en cuenta las limitaciones de aceleración del robot. Representa las velocidades posibles considerando la capacidad limitada del robot para cambiar su velocidad de manera rápida debido a sus restricciones de aceleración.

$$Vr = Vs \cap Va \cap Vd \tag{3}$$

Donde Vr es el espacio de búsqueda de velocidades óptimas, que se selecciona maximizando la siguiente función objetivo:

$$G(v,w) = \alpha * h(v,w) + \beta * d(v,w) + \gamma * v_F(v,w) \tag{4}$$

Donde "h" mide la alineación del robot con la dirección objetivo, "d" representa la distancia al obstáculo más cercano, y "vF" es la velocidad hacia adelante del robot. Los valores de α, β y γ son constantes ajustables que determinan la importancia relativa de estas tres medidas en la función objetivo. En resumen, el método DWA genera numerosas trayectorias locales en línea posibles y luego selecciona la más adecuada en función de la función objetivo. Finalmente, se ejecuta la velocidad más adecuada para seguir la trayectoria local seleccionada.

*Éxito En Alcanzar La Meta*

Este criterio generalmente se da en términos de porcentaje (%), consiste en contabilizar el porcentaje de éxitos en completar una misión de navegación respecto al total de intentos. Algunos investigadores también establecen un tiempo límite, pero suficiente para completar con éxito la misión de navegación dada, esto con el fin de descartar las pruebas en las que el robot se quede atascado navegando en un bucle interminable (McGuire et al, 2019).Un desafío adicional para los algoritmos de control es su desempeño en la navegación a través de pasajes o corredores estrechos, por lo tanto, un criterio adicional a considerar puede ser la robustez en espacios estrechos: número de pasajes estrechos atravesados con éxito.

Como ejemplo se usa en los robots de navegación que deben encontrar la salida a un laberinto, en dicha situación la triangulación de la posición se hace mediante la configuración del triángulo resuelto, donde deseamos calcular la altura (h) del triángulo con los lados 'a' y 'b' y el ángulo β. 'c' es el lado del triángulo que será desconocido, por lo que se desarrollará una fórmula que solo utilizará 'a,' 'b' y β

$$A = \frac{c.h}{2} \quad (4)$$

$$A = \frac{a.b.\sin\beta}{2} \quad (5)$$

Se aplica teorema de coseno para triangular la altura y desarrollar el triángulo reemplazando las ecuaciones anteriores.

$$h = \frac{a.b.\sin\beta}{\sqrt{a^2 + b^2 - 2.a\cos\beta}} \quad (6)$$

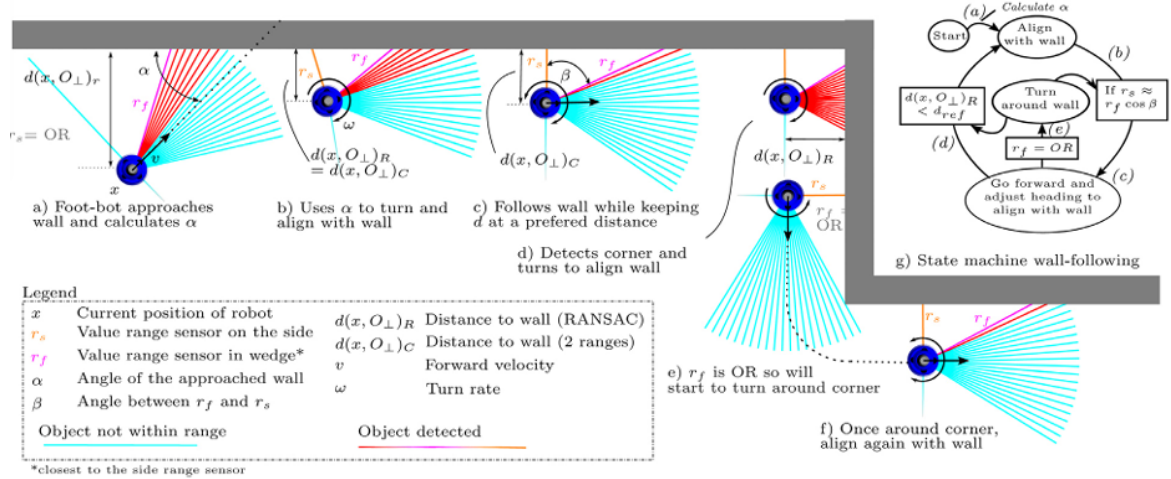

Ilustración 1 Diagrama para explicar el paradigma de seguimiento de pared para una dirección local en el lado derecho con "(g)" representando la máquina de estados correspondiente. "OR" significa "fuera de rango". Tomado de (McGuire et al, 2019).

*Tiempo En Alcanzar El Objetivo*

El tiempo de ejecución es un criterio fundamental utilizado en la mayoría de los artículos donde se comparan algoritmos. Consiste en medir el tiempo que el robot tarda en alcanzar la meta. En simulaciones de sistemas determinísticos, se obtiene el mismo resultado ante las mismas condiciones de simulación. Sin embargo, para que la simulación se acerque más a la realidad, donde existen factores como el desgaste de las baterías, la fricción entre las ruedas y el suelo, las condiciones ambientales, entre otros, se puede introducir ruido en el sistema. Esto se logra mediante la adición de perturbaciones en la lectura de los sensores o en las señales de control a los actuadores. En la simulación, se pueden utilizar generadores de números aleatorios para modelar el error o el ruido.

Otro criterio relacionado es el número de ciclos de procesamiento, que equivale al número aproximado de operaciones realizadas para completar una misión. Es importante tener en cuenta que diferentes tipos de robots pueden tener diferentes tipos de procesadores, es decir, capacidades de cómputo distintas. Esto influye en el tiempo total que el robot tarda en completar la misión y en la comparación de los algoritmos utilizando este criterio (Tai et al., 2020).

*Periodos De Control*

Los periodos de control se refieren a la cantidad de veces que el planificador toma decisiones para alcanzar el objetivo. Esta medida está relacionada con el número de iteraciones o pasos que el robot necesita para completar la misión. Si el robot se mueve a una velocidad lineal constante (v), los periodos de control proporcionan una estimación del tiempo empleado en completar la misión. Cuanto mayor sea el número de periodos de control, mayor será la cantidad de decisiones tomadas y, potencialmente, mayor será el tiempo requerido para alcanzar el objetivo.

*Criterios Basados En El Error*

En un sistema de control, el error se define como la diferencia entre la variable controlada (también conocida como variable del proceso) y el valor de referencia o set-point. En un sistema de control, el objetivo es que el error tienda a cero, lo que indica un buen desempeño del sistema.

Una forma de evaluar el desempeño de un sistema de control es cuantificar el error acumulativo. En el caso de un robot móvil, el error acumulativo proporciona una medida numérica de qué tan "bueno" es el desempeño de un controlador específico para el sistema de tracción o dirección del robot.

En controladores de tiempo discreto, es necesario conocer el error $e(nT)$ en cada instante de muestreo, donde T es el período de muestreo y fs es la frecuencia de muestreo. La frecuencia de muestreo tiene un impacto directo en la exactitud de las mediciones y en la capacidad del controlador para detectar y corregir el error a lo largo del tiempo.

Los criterios de desempeño basados en el error o en la integral del error tienen una fundamentación teórica bien establecida en los sistemas de control, y son ampliamente utilizados para evaluar y mejorar el desempeño de los controladores (Domański, 2020). Estos criterios permiten analizar el comportamiento del sistema en términos de la reducción del error y la estabilidad del control.

*Error Final*

El error final se refiere a la discrepancia entre la posición final del vehículo y el punto final de una trayectoria de referencia establecida. Se calcula midiendo la distancia entre la posición final del vehículo y el punto final deseado de la trayectoria de referencia. Esta medida es especialmente útil en vehículos submarinos robotizados, ya que permite detectar si el vehículo ha perdido la trayectoria a mitad del seguimiento o si ha llegado a una posición final incorrecta (Pérez et al., 2018).

Como ejemplo el articulo menciona que mediante una plataforma desarrollada, es posible evaluar automáticamente el rendimiento de las soluciones obtenidas utilizando una métrica específica. En este contexto, se emplea una medida típica en algoritmos de control para el seguimiento de trayectorias: el error integrado cuadrático (ISE) y el error final. El ISE se calcula como la suma de las distancias a la trayectoria ideal, que se encuentra a dos metros sobre la tubería, a lo largo del tiempo. Esta medida está inversamente relacionada con la calidad del seguimiento, ya que considera tanto el tiempo dedicado como la precisión en el seguimiento.

La velocidad de aumento es significativamente mayor cuando la distancia entre el vehículo y la trayectoria óptima aumenta, pero también castiga un seguimiento excesivamente lento de la trayectoria. Además, es esencial determinar si se ha llegado al final de la tubería. Para lograrlo, se calcula la distancia entre la posición final del vehículo y el extremo de la tubería. De esta manera, es posible identificar si el vehículo ha perdido la tubería durante la mitad del seguimiento o si ha detectado incorrectamente el final. Teniendo en consideración estas mediciones, la evaluación final se determina mediante la ecuación 7.

$$puntuaación = (1 - error^2) * (0,1 - \frac{errormedio^2}{0,1} + \frac{tiempo - ref}{100} \qquad (7)$$

Donde "error" representa el error final, "errormedio" es la media de errores a lo largo del seguimiento calculada a partir del ISE, "tiempo" indica el tiempo necesario para llevar a cabo la intervención, y "ref" se refiere a la referencia de tiempo de cada escenario, determinada en función de la distancia recorrida, los giros y los cambios de altura. El primer término de la ecuación evalúa la posición final del vehículo, sancionando aquellas situaciones en las que el vehículo se encuentra lejos del objetivo. El segundo término evalúa el error promedio a lo largo de la trayectoria. El último término es un bono que favorece los seguimientos rápidos y castiga los lentos en función de la complejidad de la ruta de la tubería.
En el diseño general de controladores, existen criterios de rendimiento comúnmente utilizados, como los índices que involucran la integral del error (Suarin et al., 2019). Estos criterios se basan en el error acumulativo y se pueden aplicar al seguimiento de trayectorias de referencia, indicando el error a lo largo de todo el recorrido entre la trayectoria de referencia y la trayectoria real seguida por el robot. Estos índices también se utilizan en el control de posición, distancia, orientación, formación de múltiples robots, entre otros (Caruntu et al., 2019) (Farias et al., 2020). Cuanto menor sea el error, mejor será la trayectoria recorrida y, en consecuencia, mejor será el algoritmo de control.

*Criterios De Seguridad*

Las investigaciones relacionadas con este tema, que abordan los criterios de desempeño en el control de robots y su seguridad en la navegación, se describen en el estudio realizado por Marvel and Bostelman en 2014. Estos criterios de desempeño se centran en la seguridad del robot mientras se desplaza a lo largo de una trayectoria determinada, teniendo en cuenta factores como la distancia entre el vehículo y los obstáculos encontrados en su camino, así como el número de colisiones que ocurren durante la navegación (Munoz et al., 2014). Estas investigaciones buscan garantizar un desplazamiento seguro y evitar posibles accidentes o daños durante la operación del robot.

La distancia media a los obstáculos durante toda la misión de navegación es otro criterio de desempeño utilizado en el control de robots. Este criterio permite evaluar qué tan cerca o lejos se encuentra el robot de los obstáculos en su entorno durante toda la trayectoria seguida. En un entorno sin obstáculos, este valor máximo será mayor, ya que el robot puede moverse libremente sin restricciones. Por otro lado, si el índice de distancia media a los obstáculos se desvía menos del valor máximo, significa que la ruta seguida por el robot transcurrió por zonas más libres de obstáculos, lo que indica un mejor desempeño en términos de evasión y navegación segura. Este criterio es importante para evitar colisiones y garantizar una trayectoria clara y libre de obstrucciones para el robot durante su misión de navegación.

La distancia media mínima a los obstáculos es otro criterio de desempeño utilizado en el control de robots para evaluar la seguridad durante una misión de navegación. Se promedia el valor mínimo de distancia medido por cada uno de los n sensores del robot. Este criterio proporciona una idea del riesgo que se ha corrido durante la misión en términos de la proximidad a los obstáculos.

En entornos sin obstáculos, donde no hay obstáculos cercanos al robot, se cumplirá que la distancia media mínima a los obstáculos será igual para todos los sensores. Esto indica que el robot ha mantenido una distancia segura y no ha estado expuesto a riesgos significativos de colisión o contacto con obstáculos.

En cambio, en entornos con obstáculos, cuanto menor sea la distancia media mínima a los obstáculos, mayor será el riesgo que se ha corrido durante la misión, ya que el robot ha estado más cerca de los obstáculos. Por lo tanto, un menor valor de este criterio indica una mayor probabilidad de colisiones o contacto con los obstáculos.

*Consumo Energético*

El consumo de energía es un aspecto clave en las aplicaciones de robots móviles, ya que influye en la autonomía del robot, es decir, en el tiempo que puede funcionar de manera óptima antes de quedarse sin energía. En los últimos años, se ha prestado especial atención a este tema (Stefek et al, 2020). Si un robot no cumple con los requisitos de consumo de energía, como la capacidad de funcionar de manera independiente y la posibilidad de recargarse, su rendimiento, tiempo de operación y autonomía se ven limitados (Heikkinen et al, 2018).

Los robots móviles dependen en gran medida de las baterías como fuente de energía, pero estas tienen una capacidad energética limitada. Como resultado, el tiempo de operación del robot suele ser corto, lo que puede ser insuficiente para tareas o misiones que requieren más tiempo y energía para completarse. Aumentar el tiempo de funcionamiento mediante el uso de más baterías o dirigiendo el robot a una estación de carga puede incrementar el costo o el tamaño del sistema, lo que puede ocasionar problemas de control. Una alternativa es mejorar la eficiencia energética del robot, reduciendo su consumo de energía (Armah et al, 2016).

En resumen, el consumo de energía es un aspecto crucial en los robots móviles, ya que afecta su autonomía y rendimiento. Es importante optimizar el diseño, los componentes y los algoritmos de control para reducir el consumo de energía y mejorar la eficiencia. Esto permitirá prolongar el tiempo de funcionamiento del robot y su capacidad de operar sin agotar rápidamente la energía, sin necesidad de aumentar significativamente el tamaño o el costo del sistema.

## Agradecimientos

Los autores agradecen a la Universidad Nacional Abierta y a Distancia por su financiación bajo el proyecto de investigación ECBTIPIE042022 Implementación de una técnica de control basada en co-diseño H/S para robots móviles basados en FPGA.